\documentclass[conference]{IEEEtran}
\ifCLASSINFOpdf
\else
\fi
\usepackage[cmex10]{amsmath}
\usepackage{mathtools}
\usepackage{amsfonts} 
\usepackage{siunitx}
\usepackage{algorithm,algorithmic}
\usepackage{booktabs}
\usepackage{threeparttable}

\usepackage{mathrsfs,amsmath}   

\newcommand{\lp}{\left(}
\newcommand{\rp}{\right)}

\DeclareSIUnit\Molar{\textsc{m}} 
\DeclareSIUnit{\pH}{pH}
\DeclareSIUnit{\pixel}{px}

\renewcommand{\vec}[1]{\text{\boldmath$#1$}} 

\makeatletter
\def\ps@IEEEtitlepagestyle{%
  \def\@oddfoot{\mycopyrightnotice}%
  \def\@evenfoot{}%
}
\def\mycopyrightnotice{%
  {\footnotesize \textbf{978-1-4673-9091-0/15/\$31.00 \copyright 2015 IEEE.}\hfill}
  \gdef\mycopyrightnotice{}
}

\begin{document}
%
\title{2D Discrete Fourier Transform with Simultaneous Edge Artifact Removal for Real-Time Applications \vspace{-0.4cm}}
\begin{author}
\author{\author{\IEEEauthorblockN{Faisal Mahmood\IEEEauthorrefmark{1}\IEEEauthorrefmark{2}, M\"art Toots\IEEEauthorrefmark{2}, Lars-G\"oran \"Ofverstedt and Ulf Skoglund\IEEEauthorrefmark{1}} \IEEEauthorblockA{Structural Cellular Biology Unit, Okinawa Institute of Science \& Technology (OIST), Okinawa, Japan.\\ Email: \{faisal.mahmood, mart.toots, lg.ofverstedt, ulf.skoglund\}@oist.jp \\ \IEEEauthorrefmark{2}Equal Contribution, \IEEEauthorrefmark{1} Corresponding Authors}}}
\end{author}

\maketitle

\begin{abstract}

Two-Dimensional (2D) Discrete Fourier Transform (DFT) is a basic and computationally intensive algorithm, with a vast variety of applications. 2D images are, in general, non-periodic, but are assumed to be periodic while calculating their DFTs. This leads to cross-shaped artifacts in the frequency domain due to spectral leakage. These artifacts can have critical consequences if the DFTs are being used for further processing. In this paper we present a novel FPGA-based design to calculate high-throughput 2D DFTs with simultaneous edge artifact removal. Standard approaches for removing these artifacts using apodization functions or mirroring, either involve removing critical frequencies or a surge in computation by increasing image size. We use a periodic-plus-smooth decomposition based artifact removal algorithm optimized for FPGA implementation, while still achieving real-time ($\ge$23 frames per second) performance for a 512$\times$512 size image stream. Our optimization approach leads to a significant decrease in external memory utilization thereby avoiding memory conflicts and simplifies the design. We have tested our design on a PXIe based Xilinx Kintex 7 FPGA system communicating with a host PC which gives us the advantage to further expand the design for industrial applications.

\end{abstract}

%
\IEEEpeerreviewmaketitle

\section{Introduction}

Discrete Fourier Transform (DFT) is a commonly used and vitally important function for a vast variety of applications including, but not limited to digital communication systems, image processing, and biomedical imaging. Fourier image analysis simplifies computations by converting complex convolution operations in the spatial domain to simple multiplications in the frequency domain.
Due to their computational complexity, DFTs often become a computational constraint for applications requiring high throughput and near real-time operations. The Cooley-Tukey Fast Fourier Transform (FFT) algorithm [1], first proposed in 1965, reduces the complexity of DFTs from $O(N^2)$ to $O(Nlog N)$ for a 1D DFT. However, in the case of 2D DFTs, 1D FFTs have to be computed in two-dimensions, increasing the complexity to $O(N^2 log N)$, thereby making 2D DFTs a significant bottleneck for real-time machine vision applications [2].

There are several resource-efficient, high-throughput implementations of 2D DFTs. Most FPGA based 2D FFT implementations rely upon repeated invocations of 1D FFTs by row and column decomposition (RCD) with efficient use of external memory [2][3][4]. Many of these achieve real-time or near real-time performance ($\ge$23 frames per second for a standard $512 \times 512$ image).

While calculating 2D DFTs it is assumed that the image is periodic, which is usually not the case. The non-periodic nature of the image leads to artifacts in the Fourier transform, usually known as edge artifacts or series termination errors. These artifacts appear as several crosses of high-amplitude coefficients in the frequency domain, as seen in [6]. Such edge artifacts can be passed to subsequent stages of processing and in biomedical applications they may lead to critical misinterpretations of results. No current 2D FFT FPGA implementation addresses this problem directly. These artifacts may be removed during pre-processing, using mirroring, windowing, zero padding or post-processing, e.g., filtering techniques; however, these techniques are usually computationally intensive and often tend to modify the transform. The most common approach is by ramping the image at corner pixels to slowly attenuate the edges. Ramping is usually accomplished by an apodization function such as a Tukey (tapered cosine) or a Hamming window, which smoothly reduces the intensity to zero. Such an approach can be implemented on an FPGA as a pre-processing operation by storing the window function in a Look-up Table (LUT) and multiplying it with the image stream before calculating the FFT [5]. Although, this approach is not extremely computationally intensive for small images, it inadvertently removes necessary information from the image.  Loss of this information may have serious consequences if the image is being further processed with several other images to reconstruct a final image that is used for diagnostics or other decision-critical applications. Another common method is by mirroring the image from $N \times N$ to $2N \times 2N$. Doing so makes the image periodic, thereby removing edge artifacts. However, this not only increases the size of the image by 4x, but also makes the transform symmetric, which generates an inaccurate phase component. 

Most RCD-based 2D FFT FPGA implementations have two major design challenges: 1) The 1D FFT implementation needs to have a reasonably high-throughput and needs to be resource efficient. 2) External DRAM needs to be efficiently addressed and have a high-bandwidth because images are usually large and intermediate storage is required between row and column 1D FFT operations.

Periodic plus smooth decomposition (PSD) [6], described in section II, provides an efficient solution for edge artifact removal from 2D DFTs with minimal amputation of useful information from the image. In section III, we describe an optimization of PSD for FPGA implementation, which reduces the number of 1D FFT invocations and requires less frequent access to external DRAM. In section IV we further describe the hardware set-up and propose an architecture for optimized PSD and suggest means for generalizing it. Section V presents experimental results and conclusions.

\section{Periodic Plus Smooth Decomposition for Edge Artifact Removal}

Periodic plus smooth decomposition (PSD) involves decomposing the image into a periodic and smooth component. The smooth component is calculated from the boundary of the image and is then subtracted from the image to derive the periodic component [6]. 

Let us have discrete $n$ by $m$ gray-scale image $\vec I$ on a finite domain $\Omega = \{0, 1, \dots, n-1\} \times \{0, 1, \dots, m-1\}$. The discrete Fourier transform (DFT) of $\vec I$ is defined as

\vspace{-0.4cm}
\begin{equation}
	\vec{\hat I}(s,t) = \sum_{(i,j) \in \Omega} \vec I(i,j) \exp \lp -i2\pi \lp \frac{si}{n} + \frac{tj}{m} \rp\rp
\end{equation}
This is equivalent to a matrix multiplication $\vec {W} \vec I \vec {V}$, where
\begin{equation}
	\vec {W} = 
	\begin{pmatrix}
			1     & 1       & 1          & \dots & 1 \\
			1     & w       & w^2        & \dots & w^{n-1} \\
			1     & w^2     & w^4        & \dots & w^{2(n-1)}\\
			\dots & \dots   & \dots      & \dots & \dots \\
			1     & w^{n-2} & w^{2(n-2)} & \dots & w^{(n-2)(n-1)}\\
			1     & w^{n-1} & w^{2(n-1)} & \dots & w^{(n-1)(n-1)}		
	\end{pmatrix}
\end{equation}
and
\begin{equation}
	w^k = \exp \lp -i \frac{2 \pi}{n} \rp ^ k = \exp \lp -i \frac{2 \pi k}{n} \rp.
\end{equation}
$ \vec {V} $ has the same structure as $\vec {W}$ but is m-dimensional. Since $w^k$ has period $n$ which means that $w^k = w^{k + ln}\,,\,\forall k,l \in \mathbb{N}$ and therefore,
\vspace{-0.2cm}
\begin{equation}\label{matrix:w}
	\vec {W} = 
	\begin{pmatrix}
			1     & 1       & 1       & \dots & 1       & 1 \\
			1     & w       & w^2     & \dots & w^{n-2} & w^{n-1} \\
			1     & w^2     & w^4     & \dots & w^{n-4} & w^{n-2}\\
			\dots & \dots   & \dots   & \dots &   \dots & \dots \\
			1     & w^{n-2} & w^{n-4} & \dots & w^{4}   & w^{2}\\
			1     & w^{n-1} & w^{n-2} & \dots & w^{2}   & w^{1}		
	\end{pmatrix}
\end{equation}
Since in general $\vec I$ is not $(n,m)$-periodic, there will be high amplitude edge artifacts present in the DFT stemming from sharp discontinuities between the opposing edges of the image as shown in figure 1b. Moisan [6] proposed a decomposition of $I$ into a periodic component $\vec P$, that is periodic and captures the essence of the image with all high frequency details, and a smoothly varying background $\vec S$, that recreates the discontinuities at the borders. So, $\vec I = \vec P + \vec S$. 
Periodic plus smooth decomposition can be computed by first constructing a border image $\vec B = \vec R + \vec C$, where $\vec R$ represents the boundary discontinuities when transitioning row-wise and $\vec C$ when going column-wise
\begin{equation}
	\begin{aligned}
	\vec R(i,j) & = 
	\begin{cases}
		\vec I(n -1 -i, j) - \vec I(i,j), \quad & \,i = 0 \text{ or } i = n - 1 \\
		\vec 0, \quad & \,\text{otherwise }\\
	\end{cases}\\
	\vec C(i,j) & = 
	\begin{cases}
		\vec I(i, m -1 -j) - \vec I(i,j), \quad & j = 0 \text{ or } j = m - 1 \\
		\vec 0, \quad & \text{otherwise}\\
	\end{cases}
	\end{aligned}
\end{equation}
It is obvious that the structure of the border image $\vec B$ is simple with nonzero values only in the edges as shown below:
\begin{equation}
  \vec B = \vec R + \vec C  = 
  \begin{pmatrix}
  b_{11}    & b_{12}   & \dots & b_{1,m-1}     & b_{1m} \\
  b_{21}    & 0        & \dots & 0             & -b_{21} \\
  \dots     & \dots    & \dots & \dots         & \dots \\
  b_{n-1,1} & 0        & \dots & 0             & -b_{n-1,1} \\
  b_{n1}    & -b_{12}  & \dots & -b_{1,m-1}    & -b_{nm}
  \end{pmatrix}.
\end{equation}
The DFT of the smooth component $\vec S$ can be then found by the following formula:
\begin{equation}
	\begin{aligned}
		\vec{\hat S}(s,t) = \frac{\vec {\hat B}(s,t)}{2 \cos \frac{2 \pi s}{n} + 2 \cos \frac{2 \pi t}{m} - 4},\quad \forall (s,t) \in \Omega\backslash\{(0,0)\}.
	\end{aligned}
\end{equation}

The DFT of the image $\vec I$ with edge artifacts removed is then $\vec {\hat P} = \vec {\hat I} - \vec {\hat S}$. Figures 1c and 1d show the DFT of the smooth and periodic components, respectively.

\begin{figure}
\centering
\includegraphics[width=3in]{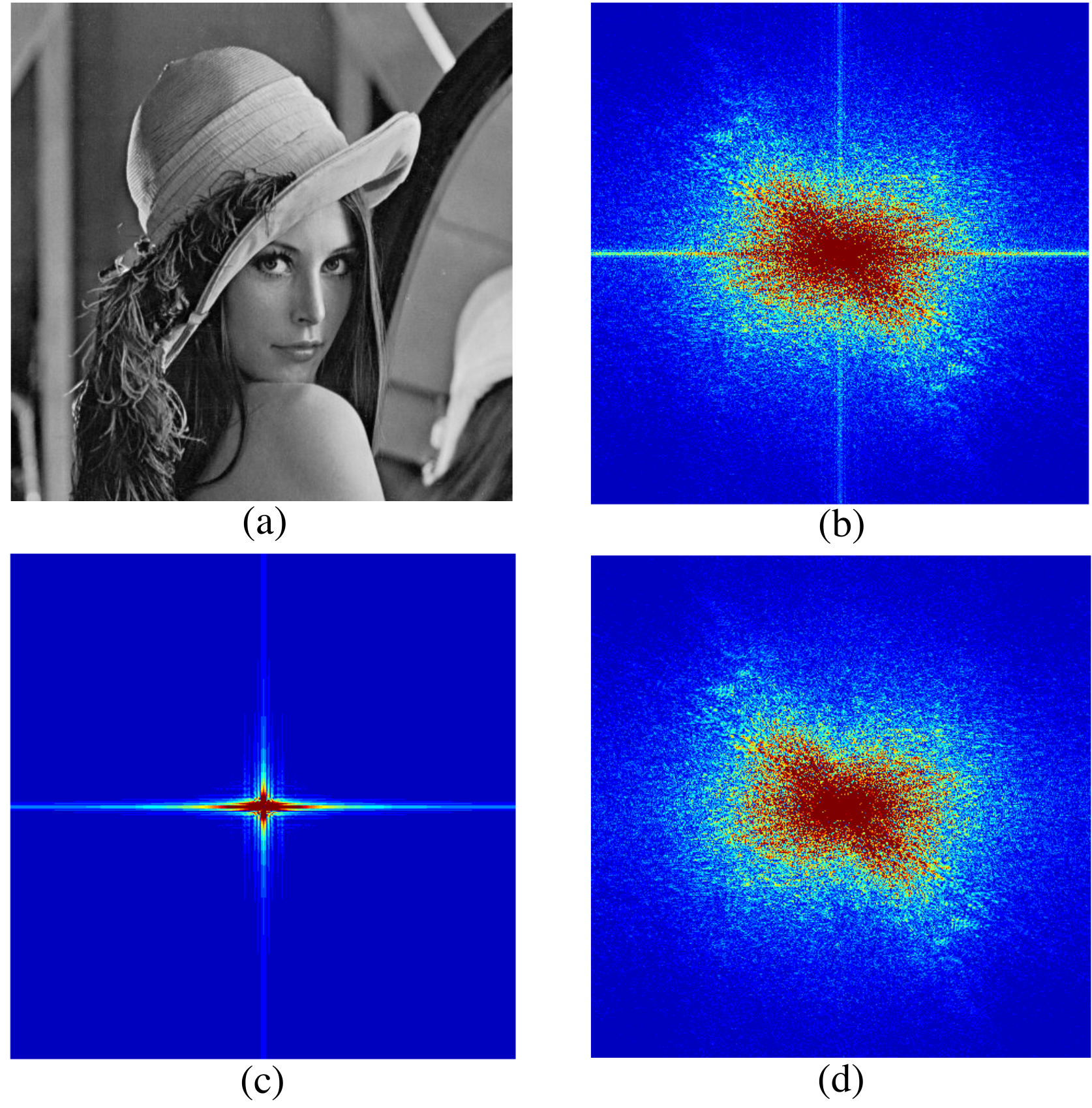}
\caption{1a) An image with non-periodic boundary. 1b) 2D DFT of 1a. 1c) DFT of the Smooth Component i.e. the removed artifacts from 1a. 1d) Periodic Component i.e DFT of 1a with Edge Artifacts removed.}
\label{fig_sim}
\end{figure}
\vspace{-0.4cm}

\section{PSD Optimization for FPGA Implementation}

In this section we optimize the original PSD algorithm so that it can be effectively configured on an FPGA. This is accomplished by using inherent symmetry between rows and columns to reduce the number of 1D FFT invocations and minimizie utilization of external DRAM. On inspecting equation (6) we realize that the boundary image $\vec B$ is symmetrical in the sense that boundary rows and columns are an algebraic negation of each other. An FFT of a column vector $\vec v$ with length $n$ is $\vec W \vec v$, where $\vec W$ is given in eq. (\ref{matrix:w}). The column-wise FFT of the matrix $\vec B$ is then
\vspace{-0.4cm}
\begin{equation}
	\vec {\hat B} = \vec W \vec B.
\end{equation}
It can be shown that the 1D FFT of the column $j \in \{2,3,\dots,m-1\}$ is
\begin{equation}
  \begin{aligned}
    \vec {\hat B_{\cdot j}} = \vec W \vec {B_{\cdot j}} =& 
    b_{1j}
    \begin{pmatrix}
    0\\
    1 - w^{n-1} \\
    1 - w^{n-2}\\
    \dots \\
    1 - w^{2} \\
    1 - w\\
  \end{pmatrix}
    = b_{1j} \vec \nu\,,
  \end{aligned}
\end{equation}
The 1D FFT of the last column $\vec {B_{\cdot m}}$ is
\begin{equation}
  \vec {\hat B_{\cdot m}} = \vec W \vec {B_{\cdot m}}
\end{equation}
\begin{equation}
  \begin{aligned}
    \vec {\hat B_{\cdot m}} =& -\vec {\hat B_{\cdot 1}} + (b_{11} + b_{1m}) \vec \nu\,.
  \end{aligned}
\end{equation}
So, the column-wise FFT of the matrix $\vec B$ is 
\begin{equation}
	\vec {\hat B} = 
	\begin{pmatrix}
		\vec {\hat B_{\cdot 1}} & b_{12} \vec \nu & \dots & b_{1,m-1} \vec \nu & -\vec {\hat B_{\cdot 1}} + (b_{11} + b_{1m}) \vec \nu \\
	\end{pmatrix}.
\end{equation}

To compute the column-by-column 1D FFT of the matrix, $\vec B$, we only have to compute the FFT of the first vector and then use the appropriately scaled vector, $\vec \nu$, to derive the remainder of the columns. The row-by-row FFT has to be calculated normally. By reducing column-by-column 1D FFT computations for the boundary image, this method can significantly reduce the number of 1D FFT invocations and reduce DRAM access for an FPGA-based implementation. This can be implemented by temporarily storing the initial vector $\vec {\hat B_{\cdot 1}}$ and scaling factors $b_{1j}$ in the block RAM/register memory, drastically reducing DRAM access and lowering the number of required 1D FFT invocations.

For a $N \times M$ image, this can reduce DRAM access from $4NM$ points to $3NM+N+M-1$ points and can reduce the number of 1D FFT invocations from to 1 column vector rather than M column vectors while calculating the column-by-column component of the 2D FFT. In other words, the number of DFT points to be computed can be reduced from $4NM$ to $3NM+M$. Table I shows a comparison of Mirroring, PSD and our proposed Optimized PSD (OPSD) with respect to DRAM access points and DFT points.  Figure 2 graphically shows that our optimized PSD method can significantly reduce reading from external memory and can reduce the overall number of DFT computations required.

\begin{figure}
\centering
\includegraphics[width=3.5in]{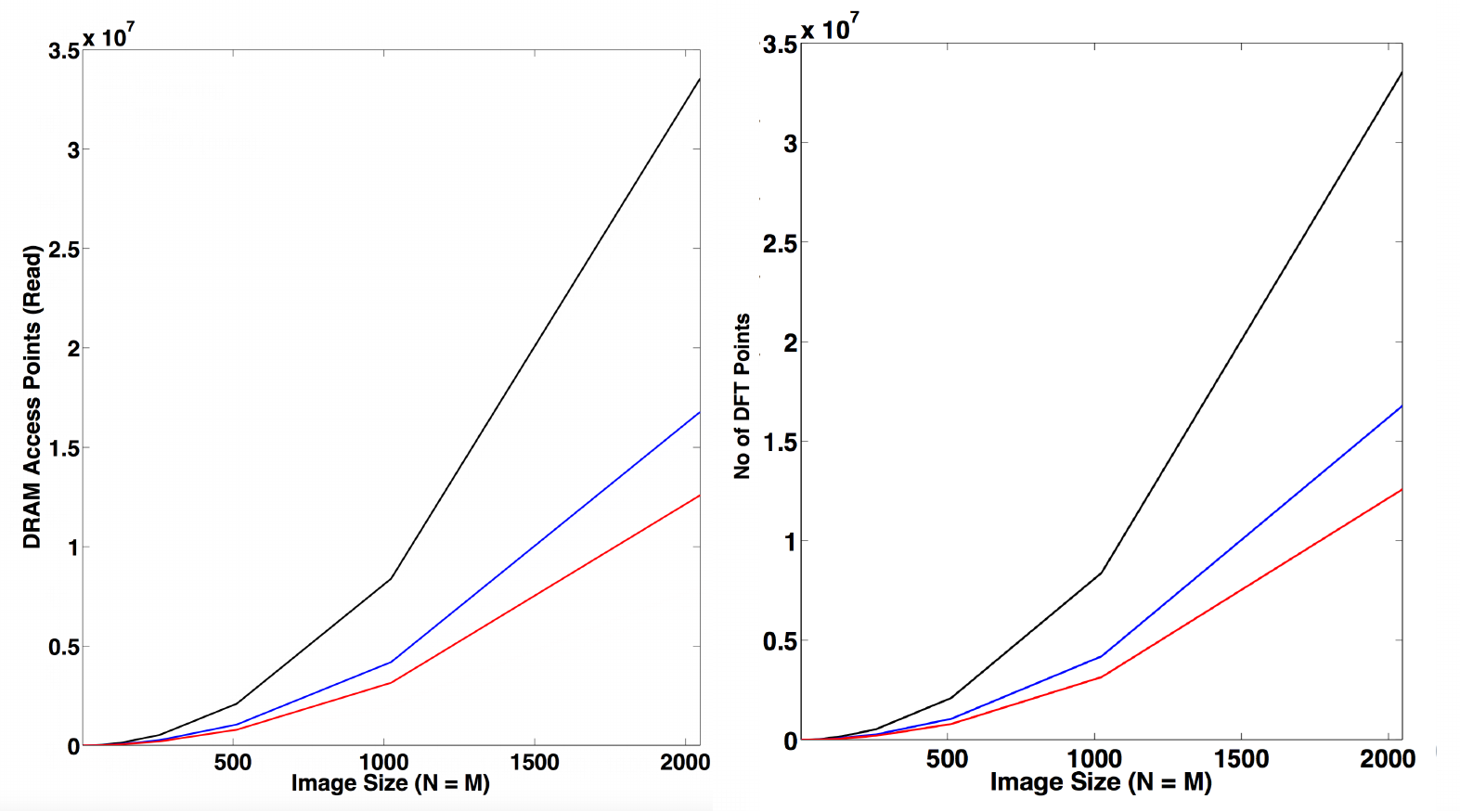}
\caption{Graphs showing DRAM access and number of DFT points to be computed with increasing image size for Mirroring (black), Periodic Plus Smooth Decomposition (blue) and our proposed Optimized Period Plus Smooth Decomposition (red).}
\label{fig_sim}
\end{figure}

\begin{table}
\caption{Comparing Mirroring, PSD and OPSD}
\begin{center}
 \begin{tabular}{lll} 
   \toprule
   \multicolumn{1}{c}{Algorithm} & {DRAM Access} & {DFT} \\
   \cmidrule(l){2-3}
      & Points & Points  \\
   \midrule
   Mirroring                      & $8NM$        & $8NM$      \\
   P+S Decomposition (PSD)        & $4NM$        & $4NM$      \\
   Optimized PSD (Proposed)       & $3NM+N+M-1$  & $3NM+M$      \\
   \bottomrule
 \end{tabular}
\label{table-tab2}
\end{center}
\end{table}

\section{FPGA Implementation of Optimized PSD}

\subsection{Hardware Configuration}

Since 2D DFTs are usually used for simplifying convolution operations in complex image processing and machine vision systems we needed to prototype our design on a system that is expandable for next levels of processing. For rapid-prototyping of our proposed optimized periodic-plus-smooth decomposition algorithm we used a PXIe (PCI eXtentions for Industry express)-based reconfigurable system. PXIe is an industrial extension of a PCI system with an enhanced bus structure that gives each connected device dedicated access to the bus with a maximum throughput of 4 GB/s. This allows a high-speed dedicated link between a host PC and several FPGAs. We used a National Instruments FlexRIO (Flexible Reconfigurable I/O) PXIe-7976R FPGA board plugged into a PXIe chassis. PXIe-7976R is equipped with a Kintex 7 FPGA and 2 GB external DRAM with data bandwidth upto 10.5 GB/s. PXIe FlexRIO FPGA boards are very adaptable and can be used to achieve high-throughput since they allow direct data transfer between multiple FPGA at rates as high as 1.5 GB/s. This can significantly simplify multi-FPGA systems, which often communicate via a host PC. This feature allows expansion of our system to further processing stages, making it flexible for a variety of applications. Figure 3 shows a basic overview of a PXIe-based, multi-FPGA system with a host PC controller connected through a high-speed bus on a PXIe chassis.

\begin{figure}
\centering
\includegraphics[width=3.2in]{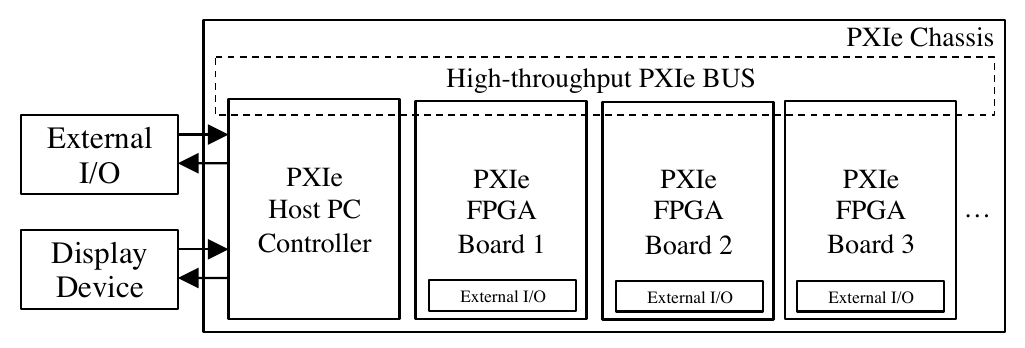}
\caption{Block diagram of a PXIe based multi-FPGA system with a host PC controller connected through a high-speed bus on a PXIe chassis.}
\label{fig_sim}
\end{figure}

\subsection{Basic Architecture} 

Most 2D FFT implementations on FPGAs use row and column decomposition (RCD) with intermediate external DRAM storage. Acceleration of RCD-based 2D FFTs is usually dependent on the throughput of the 1D FFT used for column-by-column and row-by-row 1D FFT computations. This RCD for a $N \times M$ image requires computation of $N$ row-wise and $M$ column-wise 1D FFTs. This means $MN$ (or $N^2$ if $M=N$) values must be stored after the first (row or column)-wise computations. Since 2D FFTs are usually calculated for large images, which cannot be stored on the limited embedded-block RAM, external memory must be used. Hence, acceleration also depends on bandwidth and efficient addressing of external memory. For small images block RAM or memory implemented via registers may be used as opposed to external memory. Register memory is usually faster and easier to use. Unlike external memory, it does not have limitations in terms of the number of available channels and bandwidth. However, such an approach is resource-intensive if the image is large. Uzun [3] presented an architecture for real-time 2D FFT computations using several 1D FFT processors with shared external RAM. For our 2D FFT implementation we also used an approach based on RCD with multi-core 1D FFTs. We needed a 1D FFT implementation which did not require significant resources to achieve reasonably high throughput. After comparing several 1D FFT implementations including LabView FPGA's own standard implementation, we used an Inner Loop Unrolling Technique (ILUT) [2]. A 1D FFT of length N has $logN$ FFT stages and each stage has $N/2$ butterfly units. ILUT unrolls a single FFT stage by executing several butterfly units in parallel. Figure 4 shows a basic flow of a 2D FFT implementation using ILUT. Local memory shown in Figure 4 is used to buffer data between external memory and 1D FFT cores. This local memory is divided into read and write components and is implemented using FPGA slices. This reserves block RAM (BRAM) for temporary storage of vectors required for calculating the 2D FFT of the boundary image. The Control Unit (CU) organizes scheduling of transferring data between local and external memory.

As shown mathematically in the previous section, the initial row-wise FFTs for the boundary image can be calculated by computing the 1D FFT of the first (boundary) vector and the FFTs of reaming vectors can be computed by appropriate scaling of this vector. The boundary image is calculated in the host PC. The entire boundary image does not need to be transferred to the FPGA, we only need the boundary column vector for 1D FFT calculation of the first and last column. We also need the boundary row vector for appropriate scaling of $\hat{\vec v}$ for the 1D FFT of every column between the first and last columns. To minimize data transfer between the host and the FPGA we associate an extra row and a column vector at the end of each image frame being transferred. So when transferring a $N \times M$ image frame, the number of data points sent from the host PC is $NM+N+M$. Row and column vectors of the boundary image are stored in block RAM (BRAM) while the image frame is directly stored in external DRAM. This allows column-by-column 1D FFT calculations of the boundary image to be processed in parallel with FFT computations of the actual image. A control unit schedules all read-write operations between external and local memories.

\begin{figure}
\centering
\includegraphics[width=3.3in]{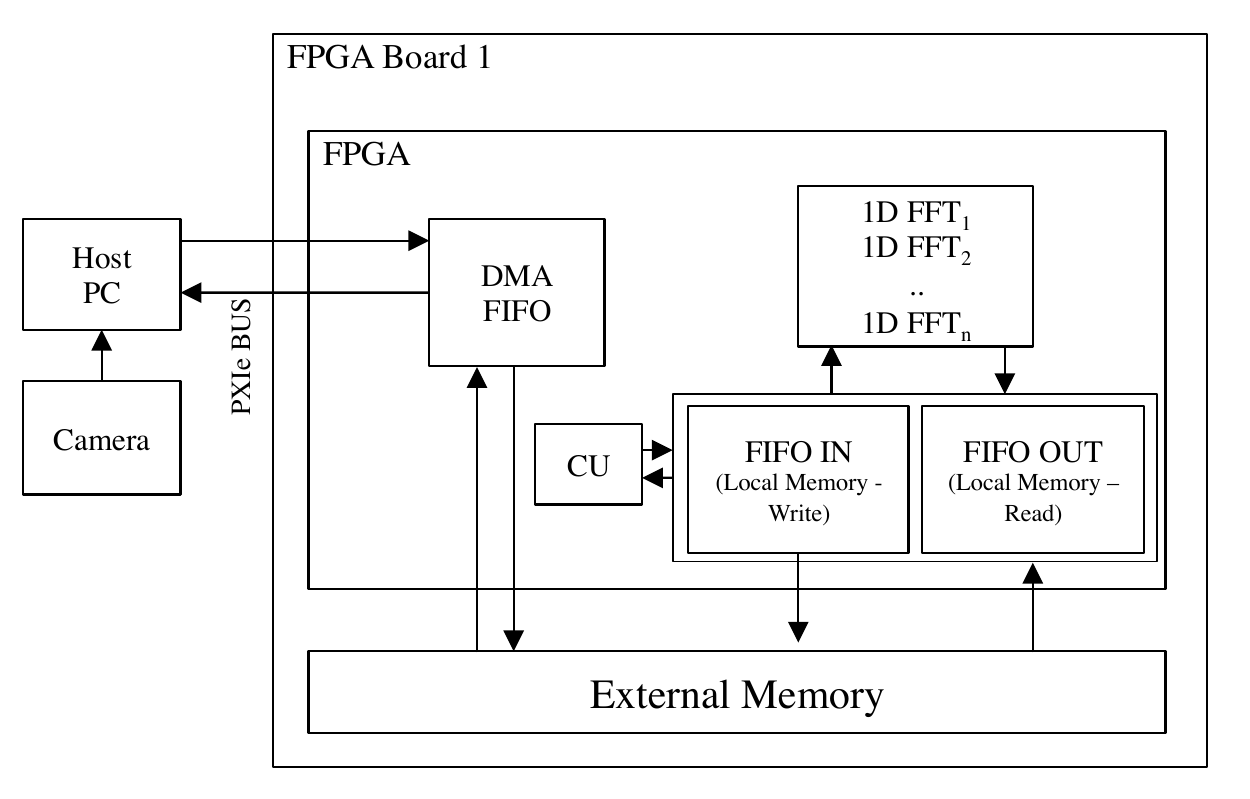}
\caption{Block diagarm of 2D FFT with ILUT showing data transfer between external memory and local memory schaduled via a Control Unit (CU)}
\label{fig_sim}
\end{figure}

\section{Results}

Table II compares several different RCD-based 2D FFT hardware implementations. None of the previous implementations use periodic-plus-smooth decomposition to simultaneously remove edge artifacts. Our implementation effectively performs twice the number of 1D FFT computations (for the original and boundary image) for each image frame, but requires only a fraction of higher run time. As demonstrated above, this acceleration has been achieved by parallelization of 2D FFT calculations for the original and boundary images and by reducing the external DRAM access by optimizing the original periodic-plus-smooth decomposition algorithm. Our methods were tested using extensive synthesis and benchmarking using a Xilinx Kintex 7 FPGA communicating with a host PC on a high-speed PXIe bus.

\section{Acknowledgements}
The authors would like to thank National Instruments Research for technical help and Shizuka Kuda for her invaluable support with logistical arrangements.

\begin{table}
  \centering
  \begin{threeparttable}[b]
  \caption{Comparison of OPSD\tnote{1} 2D FFT with regular RCD-based implementations}
  \label{tab:test2}
  \begin{tabular}{lllll}
  \toprule
  \multicolumn{1}{c}{Platform} & {SEAR \tnote{2}} & {Precision} & {Runtime} \\
   \cmidrule(l){1-5}
      & Yes/No & bits & 512x512(ms) &  1024x1024(ms) \\
         \midrule

   Kintex 7, 28nm (ours)           & Yes      & 16 (fixed)       & 32.4   & 116.7  \\
   Vertex-5, 65nm (BEE 3)[4]       & No       & 32 (single)      & 24.9   & 102.6  \\
   Vertex-E, 180nm [3]             & No       & 16 (fixed)       & 28.6   & 76.9   \\
   ASIC, 180nm [8]                & No       & 32 (single)      & 21.0   & -      \\
  \hline
  \end{tabular}
  \begin{tablenotes}
    \item[1] Optimized Periodic + Smooth Decomposition (PCD)
    \item[2] Simultaneous Edge Artifact Removal
  \end{tablenotes}
 \end{threeparttable}
\end{table}

\renewcommand{\arraystretch}{1.3}
\bibliographystyle{IEEEtran}

\begin{thebibliography}{1}

\bibitem{IEEEhowto:kopka}
Cooley, James W., and John W. Tukey. \emph{An algorithm for the machine calculation of complex Fourier series.} Mathematics of computation 19.90 (1965): 297-301.

\bibitem{IEEEhowto:kopka}
Kee, H., Bhattacharyya, S. S., Petersen, N., \& Kornerup, J. \emph{Resource-efficient acceleration of 2-dimensional Fast Fourier Transform computations on FPGAs.} Distributed Smart Cameras, 2009. ICDSC 2009. Third ACM/IEEE International Conference on. IEEE, 2009.

\bibitem{IEEEhowto:kopka}
Uzun, Isa Servan, Abbes Amira, and Ahmed Bouridane. \emph{FPGA implementations of fast Fourier transforms for real-time signal and image processing.} Vision, Image and Signal Processing, IEE Proceedings-. Vol. 152. No. 3. IET, 2005.

\bibitem{IEEEhowto:kopka}
Yu, Chi-Li, et al. \emph{Multidimensional DFT IP generator for FPGA platforms.} Circuits and Systems I: Regular Papers, IEEE Transactions on 58.4 (2011): 755-764.

\bibitem{IEEEhowto:kopka}
Bailey, Donald G. \emph{Design for embedded image processing on FPGAs.} John Wiley and Sons, 2011: 323-324.

\bibitem{IEEEhowto:kopka}
Moisan, Lionel. \emph{Periodic plus smooth image decomposition.} Journal of Mathematical Imaging and Vision 39.2 (2011): 161-179.

\bibitem{IEEEhowto:kopka}
Jung, Hyunuk, and Soonhoi Ha. \emph{Hardware synthesis from coarse-grained dataflow specification for fast HW/SW cosynthesis.} Proceedings of the 2nd IEEE/ACM/IFIP international conference on Hardware/software codesign and system synthesis. ACM, 2004.

\bibitem{IEEEhowto:kopka}
Eonic \emph{PowerFFT ASIC} [Online]. Available: http://www.eonic.com/

\end{thebibliography}
%

\end{document}